\documentclass[11pt,a4paper]{article}
\usepackage[hyperref]{emnlp2020}
\usepackage{times}
\usepackage{latexsym}

\usepackage{graphicx}
\usepackage{multirow}
\usepackage{amsmath}
\usepackage{tabularx}
\usepackage{url}
\usepackage{float}
\usepackage{xcolor}
\usepackage{tikz-qtree}
\usepackage{subfig}
\usepackage{pgfplots}
\pgfplotsset{compat=1.16} 
\usepackage{latexsym}

\usepackage{microtype}

\definecolor{orange}{RGB}{255, 230, 204}
\definecolor{yellow}{RGB}{255, 242, 204}
\definecolor{blue}{RGB}{218, 232, 252}
\definecolor{red}{RGB}{248, 206, 204}
\definecolor{green}{RGB}{213, 232, 212}

\aclfinalcopy 

\title{Unsupervised Inference of Data-Driven Discourse Structures \\using a Tree Auto-Encoder}

\author{Patrick Huber and Giuseppe Carenini\\
  Department of Computer Science \\
  University of British Columbia \\
  Vancouver, BC, Canada, V6T 1Z4 \\
  {\tt \{huberpat, carenini\}@cs.ubc.ca}}

\date{}

\begin{document}
\maketitle
\section{Introduction}
Discourse Parsing is a key NLP task for processing multi-sentential natural language. Most research in the area thereby focuses on one of the two main discourse theories -- RST \cite{mann1988rhetorical} or PDTB \cite{prasadpenn}. While these discourse theories are of great value for the field of discourse parsing and have been guiding its progress ever since, there are two major problems when data is annotated according to those theories: (1) Due to the inherently ambiguous nature of natural language, the inter-annotator agreement when following the annotation guidelines is relatively low 
\cite{carlsonbuilding} and (2) The annotation process itself is expensive and tedious, limiting the size of available gold-standard 
datasets. 

With a growing need for robust and general discourse structures in many downstream tasks and real-world applications (e.g. \citet{gerani2014abstractive, nejat2017exploring, ji2017neural}), the current lack of high-quality, high-quantity discourse trees poses a severe shortcoming. 
\begin{figure*}
    \centering
    \includegraphics[width=1\linewidth]{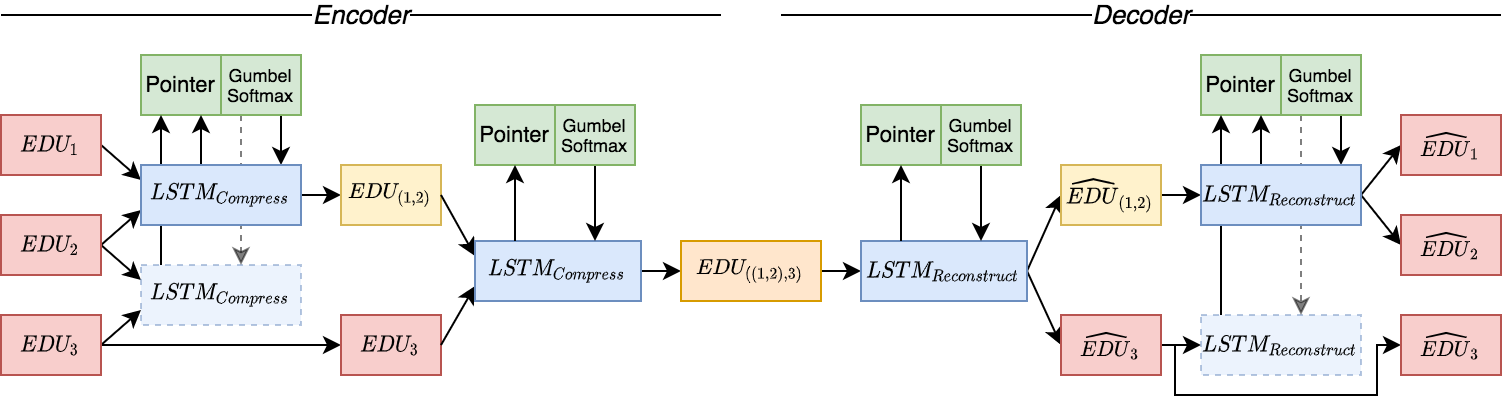}
    \caption{\textit{T-AE} (Tree Auto-Encoder) topology for unsupervised discourse parsing. Dashed components represent actions outside the computational path chosen. Inputs and outputs are dense encodings, $\widehat{EDU}_{x}$ represents reconstruction of spans, \colorbox{red}{red} = model inputs/outputs, \colorbox{blue}{blue} = TreeLSTM cells, \colorbox{green}{green} = discrete structure selector as in \citet{choi2018learning}, \colorbox{yellow}{yellow} = hidden subtree encodings, \colorbox{orange}{orange} = hidden state of the complete document.}
    \label{fig:topology}
\end{figure*}
Fortunately, there are more data-driven alternatives to infer discourse structures. For example, the recently proposed MEGA-DT treebank \cite{huber2019predicting}, whose discourse structures and nuclearity attributes are automatically inferred from large-scale \textit{sentiment} datasets, reaching state-of-the-art performance on the inter-domain discourse parsing task. Similarly, another approach by \citet{liu2019single} infers latent discourse trees from the downstream \textit{summarization} task using a transformer model.
Outside the area of discourse parsing, syntactic trees have previously been inferred according to several diverse strategies, e.g. \citet{socher2011semi, yogatama2016learning, choi2018learning, maillard2019jointly}.
In general, the approaches mentioned above 
have been shown to capture valuable information. Some models outperform baselines trained on human datasets (see \citet{huber2019predicting}), others are proven to enhance diverse downstream tasks \cite{liu2019single, choi2018learning}. However, despite these initial successes, one limitation that all aforementioned models share is the task-specificity, oftentimes only capturing downstream task-related components 
and potentially compromising the generality of the resulting trees, as for instance shown for the model using \textit{summarization} data \cite{liu2019single} in \citet{ferracane2019evaluating}. 

In order the alleviate the limitation described above, we propose a new strategy to generate tree structures in a task-agnostic, unsupervised fashion by extending a latent tree induction  framework \cite{choi2018learning} with an auto-encoding objective. 
The proposed approach can be applied to any tree-structured objective, such as syntactic parsing, discourse parsing and others. However, due to the especially difficult annotation process to generate discoursetrees, we initially intend to develop such method to complement task-specific models in generating much larger and more diverse discourse treebanks.

\section{Unsupervised Tree Auto-Encoder}
Our intuition for the tree auto-encoder comes from previous work, indicating that with available gold-standard trees, the programming-language translation task can learn valid projections in a tree-to-tree style neural network \cite{chen2018tree}. Furthermore, variational auto-encoders have been shown effective for the difficult task of grammar induction \cite{kusner2017grammar}. Generally speaking, our new model is similar in spirit to the commonly used  
sequence-to-sequence (seq2seq) architecture \cite{sutskever2014sequence}, which has also been interpreted as a sequential auto-encoder \cite{li2015hierarchical}.
However, our approach generalizes on the seq2seq model, which represents a special (left-branching) case of a general, tree-structured auto-encoder. While the sequential structure of a document 
is naturally given by the order of words, Elementary Discourse Units (EDUs) or sentences
, we face the additional difficulty to infer a valid tree structure alongside the hidden states.
To generate discrete tree structures during training, we make use of the Gumbel-Softmax decision framework \cite{jang2016categorical}.

\begin{figure}[H]
    \centering
    \includegraphics[width=.9\linewidth]{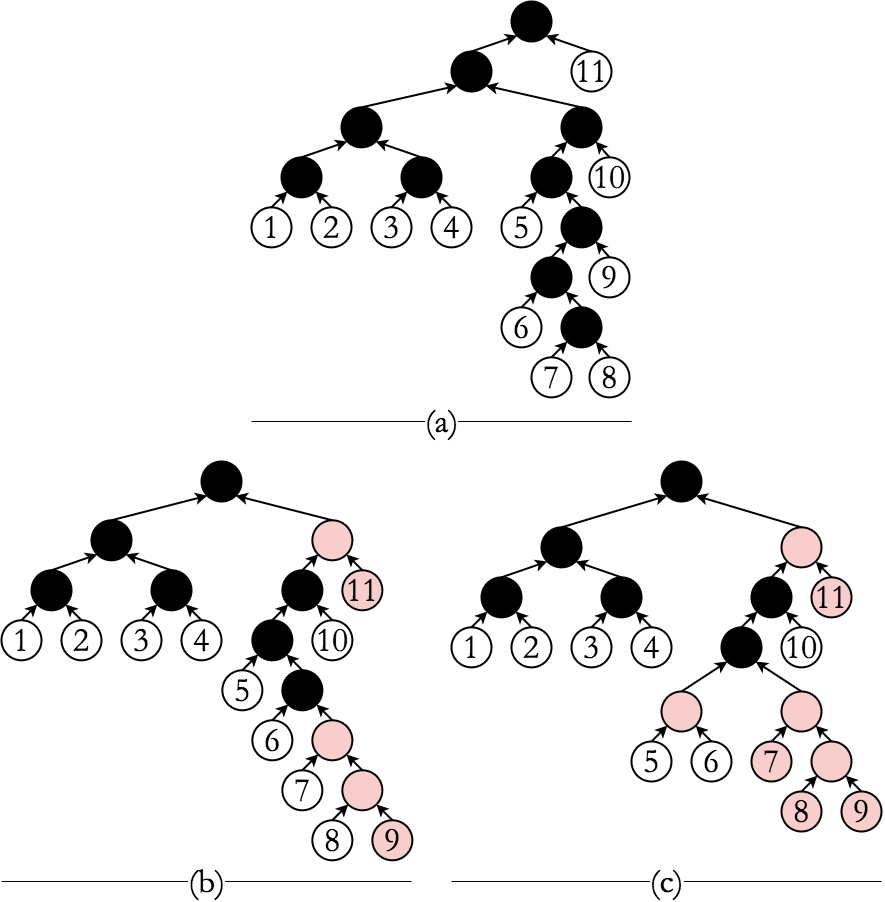}
    \caption{Discourse trees for document \textit{wsj\_1395}.\\ (a) Gold-tree, (b)/(c) T-AE outputs. Red nodes indicate deviations from the gold-tree.}
    \label{fig:example}
\end{figure}

As presented in Figure \ref{fig:topology}, the structure of our novel \textit{T-AE} (Tree Auto-Encoder) model comprises of 
an encoder and decoder component. 
The steps performed by the encoder are akin with the approach described in \citet{choi2018learning} and we direct readers to their paper for more information. Given the dense hidden state of the complete document (orange in Fig. \ref{fig:topology}), 
\citet{choi2018learning} add a multi-layer-perceptron (MLP) to predict the document-level sentiment. As a result, the obtained tree structures are mostly task-dependent \cite{williams2018latent}. With the goal to generate task-independent, more general structures, we replace the task-dependant MLP layer with our auto-encoder objective to reconstruct the original tree leaves (EDUs in our case). 
The decoder component is implemented as an inverse Tree-LSTM, recursively splitting hidden states into left and right components by applying two independent LSTM modules, guided by the predicted tree structure of the 
Gumbel-Softmax. 
This computation is recursively applied top-down until the original number of EDUs is reached. 
The EDU encodings reconstructed by the decoder are then compared with the initial EDU encodings, derived from the average GloVe embeddings of words within the EDUs. The Mean-Squared-Error (MSE) loss is computed and back-propagated to enhance the hidden states (orange/yellow in Fig. \ref{fig:topology}) and structure prediction (green in Fig. \ref{fig:topology}) of the model. To disentangle the optimization of the structure and hidden states, we apply a phased approach
. This way, the hidden states are recalculated based on the last epoch's structure prediction and vice-versa, creating a conditional back-propagation loop with a single objective in each pass over the data. We tie the decoder tree structure to the predicted encoder tree to ensure consistent compression and reconstruction, enhancing the ability to predict the leaf node encodings. 
We are currently developing and testing T-AE, with Figure \ref{fig:example} showing outputs of two independent model-runs trained on the Yelp'13 review corpus \cite{tang2015document}, compared on a randomly selected (short) example from RST-DT.
\bibliography{emnlp2020}
\bibliographystyle{acl_natbib}

\end{document}